\documentclass[letterpaper,10pt,conference]{ieeeconf}
\IEEEoverridecommandlockouts       
\let\labelindent\relax
\usepackage{amssymb}
\usepackage{amsmath}
\usepackage[english]{babel}
\usepackage{float}
\usepackage{graphicx}
\usepackage{hyperref}
\usepackage{mathtools}

\usepackage{filecontents}
\usepackage[noadjust]{cite}
\usepackage[font=footnotesize,labelfont=footnotesize]{subcaption}
\usepackage[font=footnotesize,labelfont=footnotesize]{caption}
\usepackage{comment}
\usepackage{authblk}
\usepackage{multirow}
\usepackage{xcolor}
\usepackage{algorithm}
\usepackage{algorithmicx}
\usepackage{algpseudocode}
\usepackage{xspace}
\usepackage{graphicx}

\usepackage{enumitem}

\definecolor{darkblue}{rgb}{0.15,0.15,0.55}
\definecolor{lightgrey}{rgb}{0.75,0.75,0.75}

\newcommand{\thm}{\noindent \textbf{Theorem}\xspace}

\begin{document}
\title{\LARGE \bf Merry-Go-Round: Safe Control of Decentralized Multi-Robot Systems with Deadlock Prevention\vspace{-7pt}}
\author{Wonjong Lee$^1$, Joonyeol Sim$^2$, Joonkyung Kim$^2$, Siwon Jo$^3$, Wenhao Luo$^3$, and Changjoo Nam$^{2,*}$
\thanks{This work was supported by the National Research Foundation of Korea (NRF) grants funded by the Korea government (MSIT) (No. RS-2024-00461409 and No. 2022R1C1C1008476).
$^1$Dept. of Artificial Intelligence, Sogang University. 
$^2$Dept. of Electronic Engineering, Sogang University.
$^3$Dept. of Computer Science, University of Illinois Chicago.
$^*$Corresponding author: {\tt\small cjnam@sogang.ac.kr}
}
}

\maketitle

\begin{abstract}
We propose a hybrid approach for decentralized multi-robot navigation that ensures both safety and deadlock prevention. Building on a standard control formulation, we add a lightweight deadlock prevention mechanism by forming temporary ``roundabouts'' (circular reference paths). Each robot relies only on local, peer-to-peer communication and a controller for base collision avoidance; a roundabout is generated or joined on demand to avert deadlocks. Robots in the roundabout travel in one direction until an escape condition is met, allowing them to return to goal-oriented motion. Unlike classical decentralized methods that lack explicit deadlock resolution, our roundabout maneuver ensures system-wide forward progress while preserving safety constraints. Extensive simulations and physical robot experiments show that our method consistently outperforms or matches the success and arrival rates of other decentralized control approaches, particularly in cluttered or high-density scenarios, all with minimal centralized coordination.
\end{abstract}\vspace{-3pt}

\section{Introduction}
\vspace{-3pt}
As robots have been deployed to real-world environments, navigation of multiple robots has been an important and practical problem as its applications widely range from warehouse automation to search-and-rescue operations. Multi-Agent Pathfinding (MAPF) aims to generate globally optimal and deadlock-free paths for multiple agents/robots operating in a shared environment. Various centralized algorithms, such as Conflict-Based Search (CBS)~\cite{cbs} and its numerous enhancements (e.g., \cite{ecbs,eecbs}), offer solutions that explicitly resolve conflicts to ensure all robots reach their destinations without collisions or deadlocks. 

Despite their effectiveness, conventional MAPF approaches face limitations that hinder deployment in real-world environments. Most of these methods rely on centralized computation and global replanning, requiring substantial communication bandwidth and computational resources. This becomes problematic when robots deviate from planned trajectories due to disturbances or obstacles. 
Moreover, these methods often neglect kinematic constraints, treating robots as holonomic point masses rather than considering nonholonomic motion. 
Scalability is another issue, as the state space grows exponentially with the number of robots, making real-time computation impractical. These challenges underscore the need for decentralized methods that distribute computation and enhance adaptability. 

To address these limitations, decentralized navigation strategies, including variants of Artificial Potential Fields (APF)~\cite{khatib1986real}, Control Barrier Functions (CBF)~\cite{luo2020multi,wang2017safety}, and Optimal Reciprocal Collision Avoidance (ORCA)~\cite{van2011reciprocal,bareiss2015generalized}, have gained attention for their ability to enable local decision-making without requiring a centralized planner. These methods allow robots to react to dynamic obstacles and nearby agents in real time to ensure safety. However, a key limitation of them is that they do not inherently prevent deadlocks but rather they happen to avoid them by ensuring local safety constraints. Recent works have discussed deadlock as undesired equilibrium~\cite{grover2021deadlock, reis2020control} and introduced centralized method for deadlock resolution~\cite{jankovic2023multiagent}. Unlike centralized methods that explicitly enforce deadlock-free paths through global coordination, decentralized approaches lack a systematic mechanism to prevent deadlocks in cluttered environments.

\begin{figure}[t]
\centering
\captionsetup{skip=0pt}
\includegraphics[width=0.42\textwidth]{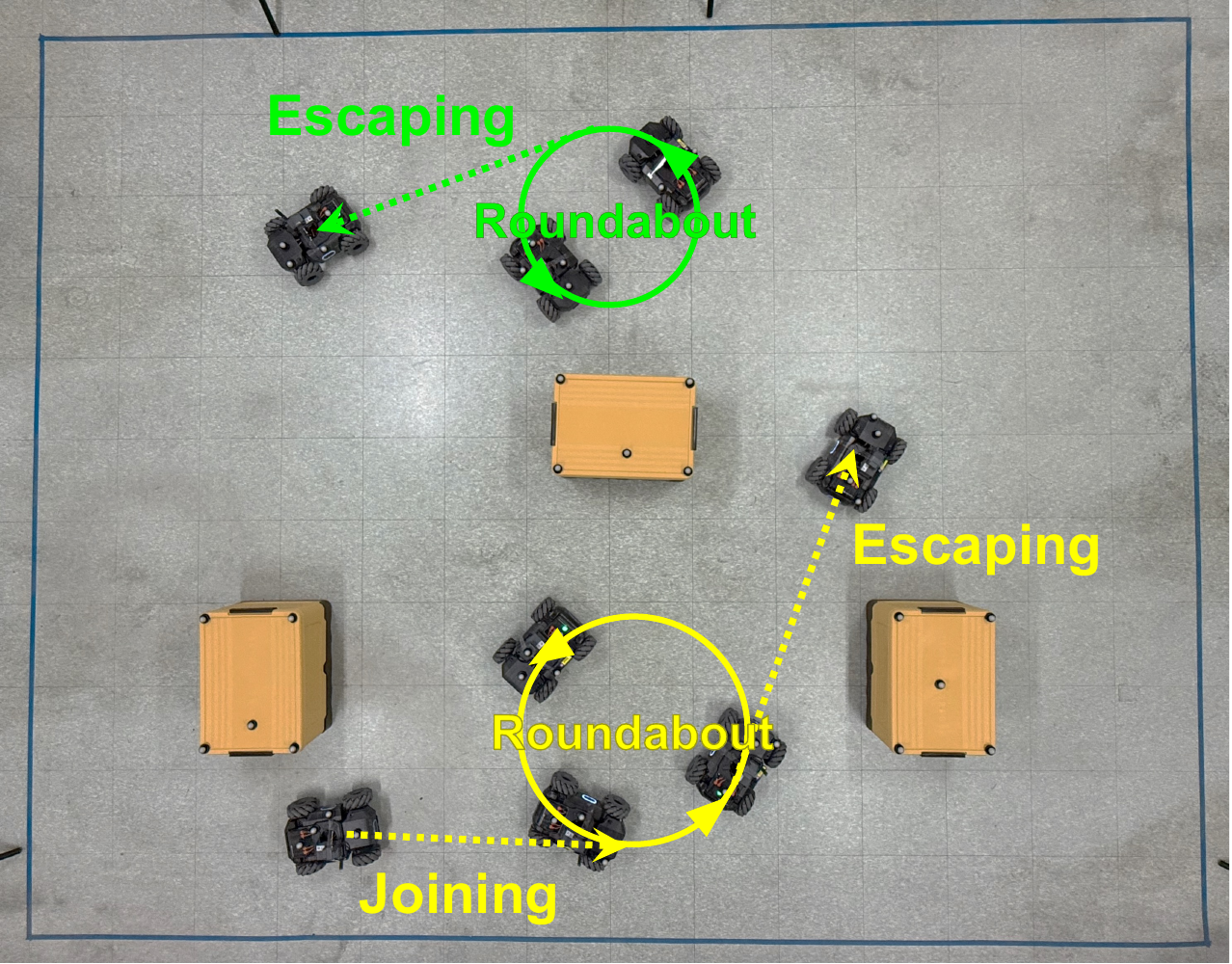}
\caption{An example where robots avoid deadlock situations using our proposed method --- Merry-Go-Round. Robots encountering a potential deadlock form a roundabout and join it until they can escape to move ahead to their goal locations.}
\label{fig:ex}\vspace{-20pt}
\end{figure}

We propose a hybrid approach that integrates decentralized reactive control with a higher-level deadlock prevention mechanism, aiming both safety and deadlock-free navigation. Our method operates by leveraging any control-based technique (e.g., \cite{wang2017safety}) to provide real-time collision avoidance while maintaining smooth motion. When a deadlock situation is predicted, a temporary roundabout maneuver is initiated among the involved robots, dynamically redirecting them until the situation is resolved.  Fig.~\ref{fig:ex} illustrates some robots that join and escape their roundabouts. The roundabout maneuver prevents deadlocks by forming a dynamic circular path where robots rotate counter-clockwise while maintaining inter-robot safety constraints.

Our method offers several key contributions to multi-robot navigation: (i) it is fully decentralized, relying solely on local peer-to-peer communication without requiring global coordination; (ii) it explicitly considers kinematic constraints, ensuring smooth motion for physical robots; (iii) it purposefully prevents deadlocks; (iv) its lightweight computation and local decision-making enable high scalability, making it effective in large-scale, dense environments; and (v) it is validated through extensive simulations and physical robot experiments, demonstrating its practicality and robustness.

\section{Related Work}
\vspace{-3pt}

MAPF has been studied as a centralized approach for computing optimal, deadlock-free paths. Algorithms like CBS~\cite{cbs} and its variants (e.g., ECBS~\cite{ecbs}, EECBS~\cite{eecbs}) resolve conflicts via conflict trees and trajectory refinements. While ensuring completeness, they rely on a central planner, incurring high computational costs and requiring global communication. Thus, they exhibit limited scalability in dynamic or partially observable environments. Additionally, MAPF solution methods often ignore kinematic constraints, causing discrepancies in real-world deployments.

For better scalability and adaptability, decentralized approaches focus on local collision avoidance. APF~\cite{khatib1986real} guides navigation with attractive and repulsive forces but struggles with local minima and oscillations. ORCA~\cite{van2011reciprocal} extends velocity obstacles for real-time collision avoidance but lacks explicit deadlock prevention. CBFs~\cite{wang2017safety} enforce safety by adjusting the controller of a robot via quadratic programming only when needed. PrSBC~\cite{luo2020multi} enhances CBFs by handling measurement uncertainty, improving robustness in noisy environments. While these methods ensure safety, they lack explicit deadlock resolution or prevention.

Recent research has introduced learning-based approaches to enhance decentralized multi-robot navigation. Reinforced Potential Fields (RPF)~\cite{zhang2023reinforced} integrate reinforcement learning with APF. Graph Control Barrier Functions (GCBF+)~\cite{zhang2025gcbf+} extend classical CBFs using graph neural networks. These approaches have demonstrated improved scalability and adaptability, particularly in highly dynamic settings. However, they still do not provide systematic deadlock prevention mechanisms, as their learned behaviors only mitigate deadlocks reactively rather than resolving them explicitly. Moreover, their generalization is limited, requiring retraining for unseen environments.

Some efforts have specifically targeted deadlock resolution in decentralized multi-robot navigation. Nonlinear Model Predictive Control (NMPC)~\cite{lafmejani2021nonlinear} formulates deadlock prevention as a constrained optimization problem, ensuring that nonholonomic robots maintain collision-free paths while avoiding deadlocks. Unlike classical reactive methods, NMPC explicitly accounts for deadlock-prone scenarios by integrating a look-ahead optimization framework that anticipates future conflicts. However, the high computational complexity of NMPC significantly limits its scalability, especially when applied to large robot teams in real-time scenarios.

Despite advancements in decentralized safety and learning-based control, the intersection of (i) decentralized execution, (ii) kinematic feasibility, (iii) systematic deadlock prevention, (iv) large-scale scalability, and (v) real-world deployment remains largely unexplored.

\section{Problem Description}
\label{sec:prob}
\vspace{-3pt}

\subsection{System Model and Formulation}
\vspace{-3pt}

We consider $N$ robots operating in a bounded 2D workspace $W \subseteq \mathbb{R}^2$. The set $O \subset W$ includes static obstacles.
For each robot $a_i$ where $i \in \{1,\ldots,N\}$, let $x_i\in \mathbb{R}^n$ and $u_i \in \mathbb{R}^m$ be its state and control input, respectively.
The system dynamics of $a_i$ follow a control-affine form:
\begin{equation}
\dot{x_i} = f_i(x_i)+g_i(x_i)u_i
\end{equation}
where $f_i: \mathbb{R}^n \rightarrow \mathbb{R}^n$ and $g_i: \mathbb{R}^n \rightarrow \mathbb{R}^{(n\times m)}$ are locally Lipschitz continuous. 
The safety of system constraints is characterized through a safe set $\mathcal{S} $. For collision avoidance between robot pairs $(a_i, a_j)$ where $i\neq j$, we define the pairwise safety function:
\begin{equation}
h(x_i,x_j)=||p_i-p_j||^2-d_{\text{safe}}^2
\end{equation}
where $p_i\in \mathbb{R}^2$ represents the position components extracted from the full state $x_i$, which may include additional state variables such as orientation. A positive scalar value $d_\text{safe}$ is the minimum allowed inter-robot distance. We set $d_\text{safe} = 2 r_\text{safe}$ where $r_\text{safe}$ is a safety margin for each robot. The global safety function is defined as:
\begin{equation}
\mathcal{S} = \{\, x_i \in \mathbb{R}^n\mid\, h (x_i,x_j) \ge 0, \forall j\neq i\}.
\end{equation}

\subsection{Control Lyapunov Functions}
To stabilize the nonlinear system of robots, we employ Control Lyapunov Functions (CLFs). A continuously differentiable function $V: D\rightarrow\mathbb{R}_{\geq0}$ is termed a CLF if it is positive definite and satisfies:
\begin{equation}
    \underset{u}{\inf} [L_{f_i} V(x) + L_{g_i} V(x) u] \leq -\gamma(V(x)) \label{eq:clf}
\end{equation}
where $\gamma:\mathbb{R}_{\geq0} \rightarrow \mathbb{R}_{\geq0}$ is a class $\mathcal{K}$ function, meaning it satisfies $\gamma(0)=0$ and $\gamma$ is strictly increasing. This definition yields the following set of stabilizing controls:
\begin{equation}
    K_{\text{clf}}(x)=\{u:L_{f_i}V(x) +L_{g_i} V(x) u \leq -\gamma (V(x))\}.
\end{equation}
If $V$ is a valid CLF, any locally Lipschitz continuous feedback $u=k(x)$ selected from $K_\text{clf}(x)$ will asymptotically stabilize the system to the equilibrium $x^*$ such that $V(x^*)=0$, typically at $x^* =0$.

\subsection{Control Barrier Functions}
In the multi-robot system described above, stabilization focuses on driving the state of each robot to a desired configuration, while safety ensures that the system state remains within the safe set $\mathcal{S}$. A continuously differentiable function $h:X\rightarrow\mathbb{R}$ is termed a Control Barrier Function (CBF) if there exists an extended class $\mathcal{K}_\infty$ function $\alpha:\mathbb{R}\rightarrow\mathbb{R}$ (i.e., strictly increasing with $\alpha(0)=0$) such that:
\begin{equation}
\begin{split}
    \underset{u}\sup[L_{f_i} h(x_i, x_j) + L_{g_i} h(x_i, x_j)u]
    \\\geq -\alpha(h(x_i,x_j)), 
    \forall j\neq i.
     \label{eq:cbf}
\end{split}
\end{equation}
The set of all safety-preserving controls can be expressed as:
\begin{equation}
\begin{split}
    K_{\text{cbf}}(x_i,x_j)=\{u: L_{f_i} h(x_i,x_j) + L_{g_i} h(x_i,x_j)u \\
    \geq -\alpha(h(x_i,x_j)), \forall j \neq i \}.
\end{split}
\end{equation}

\subsection{QP-based Control Formulation}

To synthesize controllers that ensure both stability and safety, we formulate a Quadratic Program (QP) integrating the CLFs and CBFs, following the approach presented in~\cite{ames2017control}. Let $\mathbf{u}=(u,\delta)\in\mathbb{R}^m\times\mathbb{R}$ denote our decision variable where $\delta$ is a scalar relaxation variable. The QP formulation is
\begin{align}
    &\text{min } \frac{1}{2}\mathbf{u}^TH(x_i)\mathbf{u} + F(x_i)^T\mathbf{u} \label{eq:qp_obj} \\
    &\text{s.t.} \quad L_{f_i}V(x_i) + L_{g_i}V(x_i)u + \gamma(V(x_i)) \leq \delta & \tag{CLF} \label{eq:clf_}  \\
    &\phantom{\text{s.t.}} \quad \begin{gathered}
        L_{f_i}h(x_i,x_j) + L_{g_i}h(x_i,x_j)u \\
        +\alpha(h(x_i,x_j)) \geq 0 , \forall j\neq i
    \end{gathered} & \tag{CBF} \label{eq:cbf_} 
\end{align}
where \( H(x_i) \in \mathbb{R}^{(m+1)\times(m+1)}\) is a positive definite matrix ensuring the convexity of the QP, and $F(x_i) \in \mathbb{R}^{m+1}$ represents the linear term in the objective function. 
With \( \delta \), the stability constraint is allowed to be violated to maintain safety whenever necessary. 

However, as noted in~\cite{reis2020control}, even QP-based frameworks can admit undesirable equilibrium points that cause the system to remain stuck, resulting in a deadlock and preventing the robot from reaching its goal. To address this, we propose a deadlock prevention mechanism, which uses a roundabout maneuver, to escape such equilibria.

\subsection{Assumptions}
\vspace{-3pt}

We assume all robots are homogeneous, sharing the same unicycle kinematics and communicating perfectly within $\delta_{\text{comm}}$. Without loss of generality, their sensing range $\delta_\text{sensing}$ equals the communication range. Each robot knows the environment map, its own state, and its goal position. Owing to the barrier constraints, robots cannot approach obstacles within $2r_{\text{safe}}$. Consequently, goal positions must lie at least $2r_{\text{safe}}$ from obstacles, and likewise, any two goals must be separated by more than $2r_{\text{safe}}$, ensuring each pair of robots can converge to their respective goals without violating the barrier constraints. For handling obstacles, we use a right-hand rule which makes robots encountering static obstacles move clockwise around obstacles.

\section{Merry-Go-Round for Deadlock Prevention}
We propose the Merry-Go-Round (MGR) algorithm for multi-robot navigation while ensuring safety and preventing deadlocks. We first describe the algorithm and then provide a time complexity analysis.

\subsection{Algorithm Description}
The MGR algorithm consists of (i) deadlock prediction, (ii) avoidance circle (i.e., \textit{roundabout}) generation, and (iii) escape condition checking that enables robots to safely exit the avoidance maneuver. The complete procedure is outlined in Alg.~\ref{alg:MGR}.

\setlength{\textfloatsep}{5pt}
\begin{algorithm}
{\small
    \caption{\textsc{Merry-Go-Round}}  
    \label{alg:MGR}
    \begin{algorithmic}[1]
    \Statex \hspace{-1.5em} \textbf{Input: }{Robot $a_i$ and its current position $x_i$, velocity $v_i$, and goal position $g_i$, a set of robots $a_j \in \mathcal{A}$ within $\delta_\text{comm}$ of $a_i$ ($i \neq j$), time horizon $T$, the set of roundabouts  $\mathcal{C}$, static obstacles $O$}
        \If{\textsc{receiveMGR}$(C)$}~\label{alg1:recieveMGR}
            \State $a_i.mode \leftarrow $ \textsc{joinMGR$(a_i, C)$}~\label{alg1:join_recieveMGR}
        \Else
            \For {each robot $a_j \in \mathcal{A}$}\label{alg1:for}
                \State $\{x_j, v_j, g_j\} \gets \textsc{receiveState}(a_j)$~\label{alg1:recievestate}
                \If {\textsc{isDeadlockCandidate}$(x_i,v_i,x_j,v_j,T)$}~\label{alg1:deadlock}
                    \If{\textsc{isGoalChecking}$(a_i, a_j, x_i, x_j, g_i, g_j)$}~\label{alg1:isGoalChecking}
                    \State \textbf{continue}
                    \EndIf
                    \State $\bf{c}$ $\gets \textsc{findCenter}(a_i, a_j)$~\label{alg1:findIntersection}
                    \If{$\exists C \in \mathcal{C}$ within $\delta_c$ of $\bf{c}$} ~\label{alg1:ifUpdateMGR}
                        \If {\textbf{not} \textsc{isMGRValid($C, O$)}}~\label{alg1:isMGRValid1}
                        \State $C \gets \textsc{adjustMGR}(C, O)$~\label{alg1:updateMGR}
                        \EndIf                        
                        \State $a_i.mode \leftarrow $ \textsc{joinMGR$(a_i, C)$}~\label{alg1:joinMGR}
                        \State \textsc{sendMGR}$(a_j, C)$ ~\label{alg1:sendMGR}
                    \Else
                        \State $C \gets$ \textsc{createMGR(\textbf{c})
                        }~\label{alg1:createMGR}
                        \If {\textbf{not} \textsc{isMGRValid($C, O$)}}~\label{alg1:isMGRValid2}
                            \State $C \gets \textsc{adjustMGR}(C, O)$~\label{alg1:updateMGR2}
                        \EndIf
                        \State $a_i.mode \leftarrow $ \textsc{joinMGR$(a_i, C)$}~\label{alg1:joinMGR2}
                        \State \textsc{sendMGR}$(a_j, C)$~\label{alg1:sendMGR2}
                    \EndIf
                \EndIf
            \EndFor\label{alg1:endfor}
        \EndIf
        \If{$\textsc{isEscapable}(a_i)$ \textbf{and} $a_i.mode=\textsf{MGR}$}~\label{alg1:isEscapableandaiMGR}
            \State $a_i.mode \leftarrow $ \textsc{EscapeMGR$(a_i)$}~\label{alg1:escape}        
        \EndIf
    \end{algorithmic}
}
\end{algorithm}

Alg.~\ref{alg:MGR} runs on $a_i$ in a decentralized manner in each time a control input is calculated. It receives information regarding $a_i$ (current position $x_i$, goal position $g_i$, and velocity $v_i$), nearby robots  $a_j \in \mathcal{A}$, and other information (time horizon $T$ for deadlock prediction, the set of roundabouts $\mathcal{C}$, and static obstacles $O$).

\begin{figure}[t]
\captionsetup{skip=0pt}
    \centering
    \begin{subfigure}[b]{0.62\columnwidth}
    \captionsetup{skip=0pt}
        \centering
        \includegraphics[width=\columnwidth]{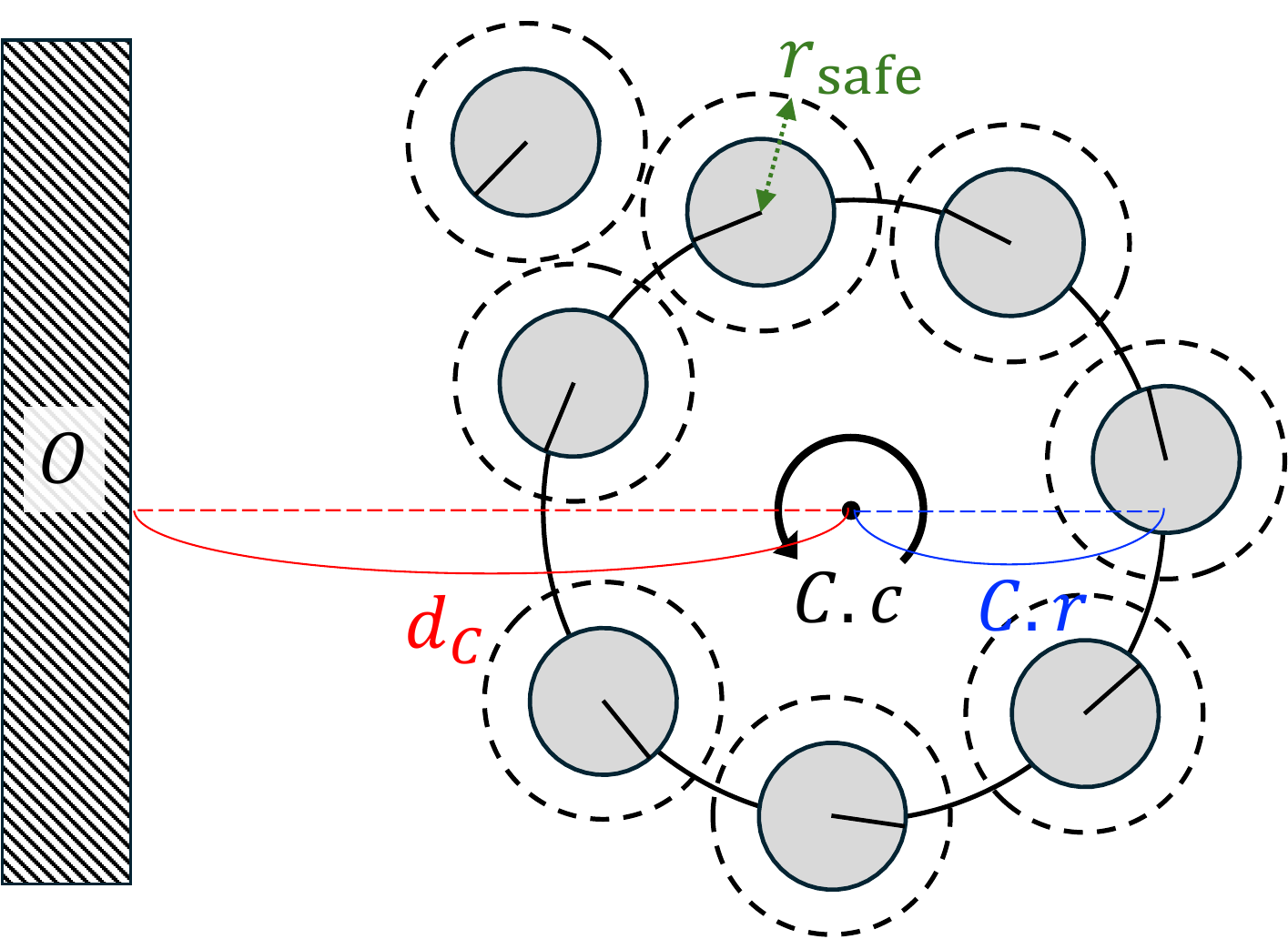}
        \caption{Valid roundabout}
        \label{fig:roundabout_1}
    \end{subfigure}
    \hfill
    \begin{subfigure}[b]{0.34\columnwidth}
    \captionsetup{skip=0pt}
        \centering
        \includegraphics[width=\columnwidth]{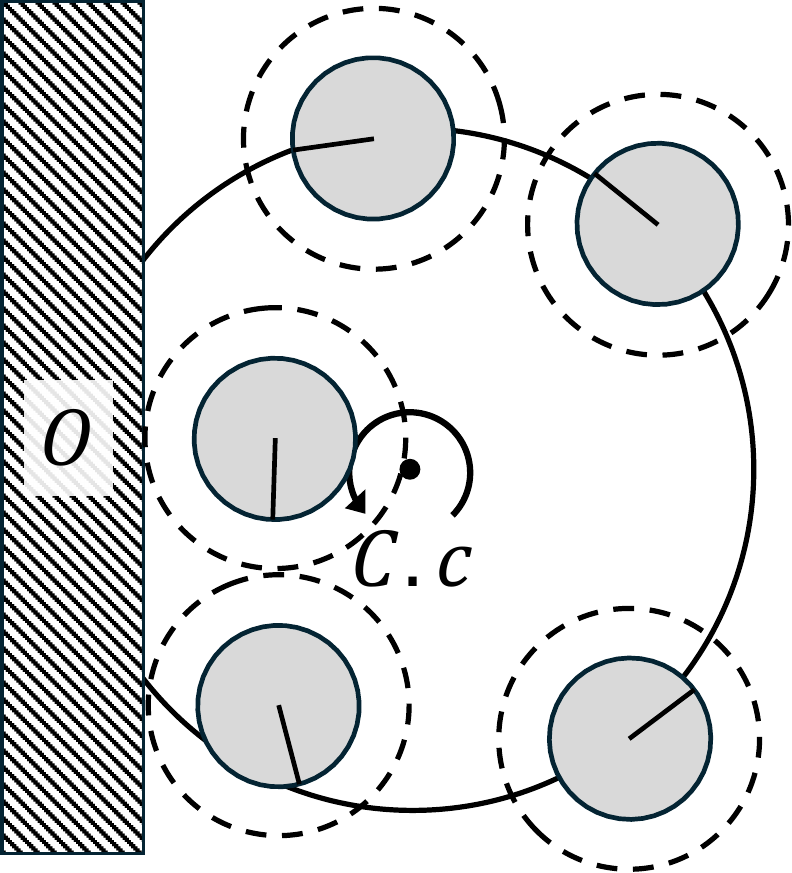}
        \caption{Invalid roundabout}
        \label{fig:roundabout_2}
    \end{subfigure}
    \caption{Roundabouts for deadlock prevention. (a) Robots, each with a safe margin $r_\text{safe}$, form a roundabout. If the roundabout reaches capacity, additional robots orbit outside. The distance $d_C$ ensures that all robots, including those on the outer orbits, remain at least $C.r$ away from the center $C.c$. (b) A roundabout can become invalid if $d_C$ is too small, since the robots can no longer maintain the required distance $C.r$ from $C.c$.}
    \label{fig:roundabout}
    \vspace{-5pt}
\end{figure}

A roundabout $C$ is a circular reference path for robots facing potential deadlock. Each robot travels around $C$ in the same direction (counterclockwise in our implementation) until it reaches an escape condition. Ideally, each robot maintains a distance of $C.r$ from the roundabout center $C.c$. However, if $C$ becomes full and cannot accommodate more robots at radius $C.r$, additional robots orbit outside while still respecting the barrier constraints, as shown in Fig.~\ref{fig:roundabout_1}. Robots in outer orbits may later move into the inner orbit if there is sufficient space without violating the barrier constraints. As described in Fig.~\ref{fig:roundabout_2}, the robots may not be able to follow $C$ if $C$ is not sufficiently apart from $O$ as the robots must ensure safety.

As a result of executing Alg.~\ref{alg:MGR}, each robot $a_i$ sets its navigation mode $a_i.mode$ to either \textsf{GOAL} (moving directly to its goal) or \textsf{MGR} (following a roundabout). The controller then applies different strategies depending on the mode:

\begin{itemize}
\item \textsf{GOAL} mode: The nominal control input for $a_i$ is computed based on feedback control using position error to guide the robot toward its goal position. The control law generates velocity commands in single integrator space, which are then converted to unicycle control inputs through state-dependent transformations in~\cite{pickem2017robotarium}.

\item \textsf{MGR} mode: The desired velocity $\mathbf{v}_i^{\text{des}}$ for $a_i$ is computed as the sum of a tangential velocity component $\mathbf{v}_i^{\text{tan}}$ that drives counterclockwise circular motion and a radial velocity component $\mathbf{v}_i^{\text{rad}}$ that maintains the desired radius from $C.c$: 
\begin{equation}
\mathbf{v}_i^{\text{des}} = 
\frac{\mathbf{v}_i^{\text{rad}}+\mathbf{v}_i^{\text{tan}}}
{\| \mathbf{v}_i^{\text{rad}} +\mathbf{v}_i^{\text{tan}}\|} v_\text{max}~\label{eq:mgr}
\end{equation}
where $\mathbf{v}_i^{\text{tan}} = v_{\text{max}} \begin{bmatrix} -\sin\theta_i \ \cos\theta_i \end{bmatrix}$
with $\theta_i = \tan^{-1}[({y_i - c_y})/({x_i - c_x})] $ is the angle from $C.c$ to $a_i$.
Additionally, $\mathbf{v}_i^{\text{rad}} = \frac{k_p}{C.n}(\|x_i - C.c\|-C.r) \frac{C.c - x_i}{\|C.c-x_i\|}v_\text{max}$
where $0<k_p\leq1$ is a proportional gain constant and $C.n$ is the number of robots current taking $C$. 
\end{itemize}

In Alg.~\ref{alg:MGR}, $a_i$ checks if any roundabout information $C$ has been received from another robot (line~\ref{alg1:recieveMGR}), where $C$ contains its center position $C.c$, radius $C.r$, and the number of robots currently taking $C$, indicated by $C.n$. If received, $a_i$ immediately joins $C$ (line~\ref{alg1:join_recieveMGR}) and the navigation mode of $a_i$ remains \textsf{MGR} or switches to \textsf{GOAL} depending on the previous mode. 

\begin{figure}[t]
    \captionsetup{skip=0pt}
    \centering
    \begin{subfigure}[b]{0.405\columnwidth}
    \captionsetup{skip=0pt}
        \centering
        \includegraphics[width=\columnwidth]{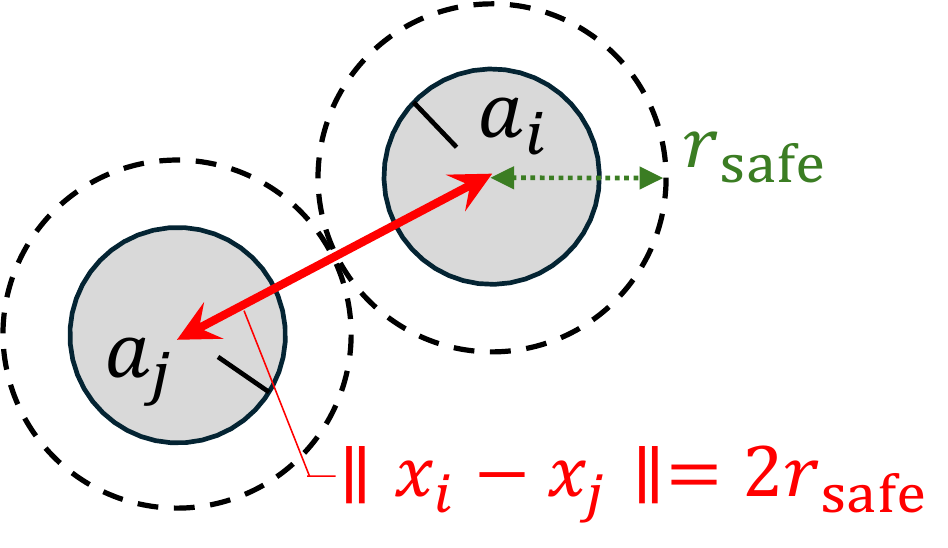}
        \caption{Condition (\ref{eq:condition_1})}
        \label{fig:condition_1}
    \end{subfigure}%
    \begin{subfigure}[b]{0.585\columnwidth}
    \captionsetup{skip=0pt}
        \centering
        \includegraphics[width=\columnwidth]{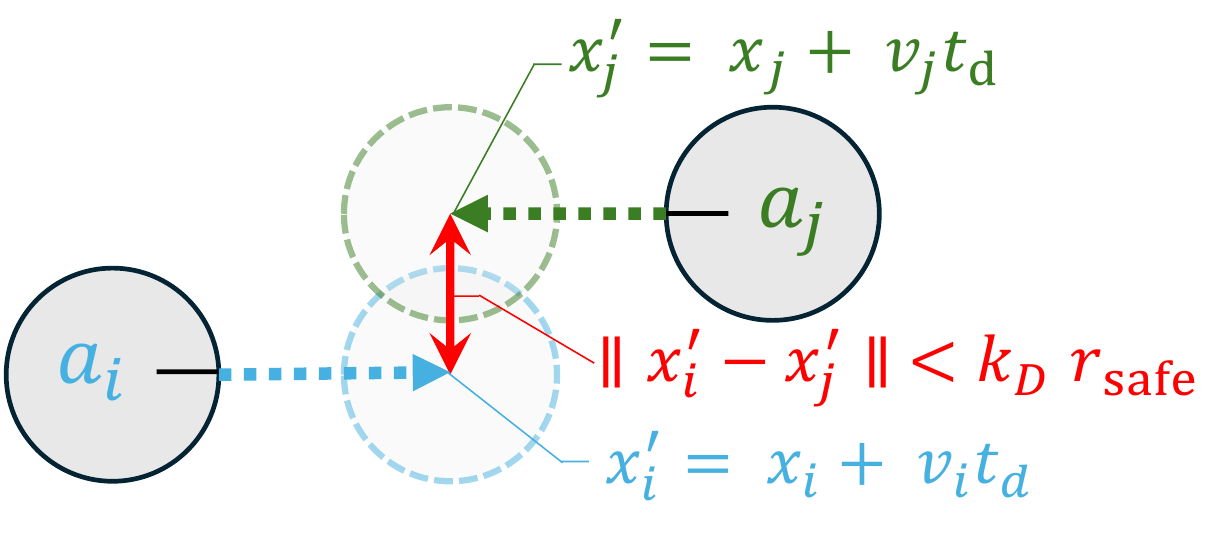}
        \caption{Condition (\ref{eq:condition_2})}
        \label{fig:condition_2}
    \end{subfigure}
    \caption{Deadlock conditions for $a_i$ and $a_j$. (a) Robots are at the barrier constraint distance $\|x_i - x_j\| = 2r_\text{safe}$. (b) There exists a time $t_d \in[0,T]$ where the estimated distance between $a_i$ and $a_j$ violates the barrier constraints.}
    \label{fig:condition}
    \vspace{-5pt}
\end{figure}

If no message has been received, Alg.~\ref{alg:MGR} checks if there is a potential deadlock between $a_i$ and $a_j$ (line~\ref{alg1:deadlock}) according to two conditions:
\begin{enumerate}[label=\alph*)]
\item The distance between $a_i$ and $a_j$ is $2r_{\text{safe}}$ (e.g., Fig.~\ref{fig:condition_1}):
\begin{equation}
    \|x_i-x_j\| =2r_{\text{safe}}~\label{eq:condition_1}
\end{equation}
which indicates that robots are already in barrier constraint active sets. In this case, the robots would be in an undesirable equilibrium, possibly causing a deadlock owing to their safety constraints being active, as proven in~\cite{reis2020control}.

\item The estimated trajectories of $a_i$ and $a_j$ are within $k_D r_{\text{safe}}$ over a time horizon $T$ (e.g., Fig.~\ref{fig:condition_2}):
\begin{equation}
    \min \| x_i^\prime - x_j^\prime\| \leq k_Dr_{\text{safe}}~\label{eq:condition_2}
\end{equation}
where $x_i^\prime = x_i + v_i t_d$, $x_j^\prime = x_j + v_j t_d$, and $0 \leq t_d \leq T$. The range of a constant $k_D$ is $1\leq k_D < 2$ because robots could enter the barrier constraint active sets of each other in this range, potentially leading to a deadlock situation.
\end{enumerate}

In line~\ref{alg1:isGoalChecking}, \textsc{isGoalChecking} checks if $a_i$ and $a_j$ are already in close proximity (within $\epsilon < r_\text{safe}$) to their respective goal positions so they can reach their goals without a deadlock. If this condition is satisfied, the deadlock prevention for $a_i$ and $a_j$ is skipped. Specifically, the condition
\begin{equation}
\begin{aligned}
\|x_i - g_i\| \leq \epsilon \text{ and } \|x_j - g_j\| &\leq \epsilon \\
\text{and } a_i.mode = a_j.mode &= \textsf{GOAL}
\end{aligned}
\end{equation}
describes a situation where both robots are about to reach their destinations and not currently engaged in any roundabout maneuver.

If a potential deadlock is detected between $a_i$ and $a_j$, \textsc{findCenter} finds $\textbf{c}$ which will be used as the initial center of a roundabout for the robots (line~\ref{alg1:findIntersection}). We use the midpoint between $a_i$ and $a_j$ for simplicity, which can be calculated more accurately by predicting the intersection of the actual trajectories of the robots. 

If there exists a roundabout $C$ within proximity $\delta_c$ of $\mathbf{c}$ (line~\ref{alg1:ifUpdateMGR}), $a_i$ and $a_j$ join $C$ without generating a new roundabout. However, we must then verify that $C$ is sufficiently distanced from $O$ so that any robots using $C$ can still navigate around $C.c$, even if $C$ is already at capacity and additional robots must orbit on larger radii, as illustrated in Fig.~\ref{fig:roundabout_1}. In line~\ref{alg1:isMGRValid1}, \textsc{isMGRValid} confirms $d_C \ge C.r + kC.n$, ensuring there is enough clearance from $O$ to accommodate the outer orbits.
If $C$ is invalid, \textsc{adjustMGR} finds a new center of $C$ (line~\ref{alg1:updateMGR}) such that $C.c$ and the closest obstacle in $O$ is at least $d_C$. In our implementation, \textsc{adjustMGR} discretizes a region of the environment around the current $C.c$ and searches for a valid location for a new $C.c$. It constructs a grid and marks each cell as either \emph{valid} or \emph{invalid}. A cell is marked \emph{valid} only if it is at least $C.r + kC.n$ away from every obstacle. The valid cell with the lowest index among those having the closest distance to the current $C.c$ is selected as the new $C.c$.\footnote{An implementation of \textsc{adjustMGR} can vary. A sampling-based method also can efficiently find a new $C.c$.}

\begin{figure}[t]
\captionsetup{skip=0pt}
\centering
\includegraphics[width=0.5\columnwidth]{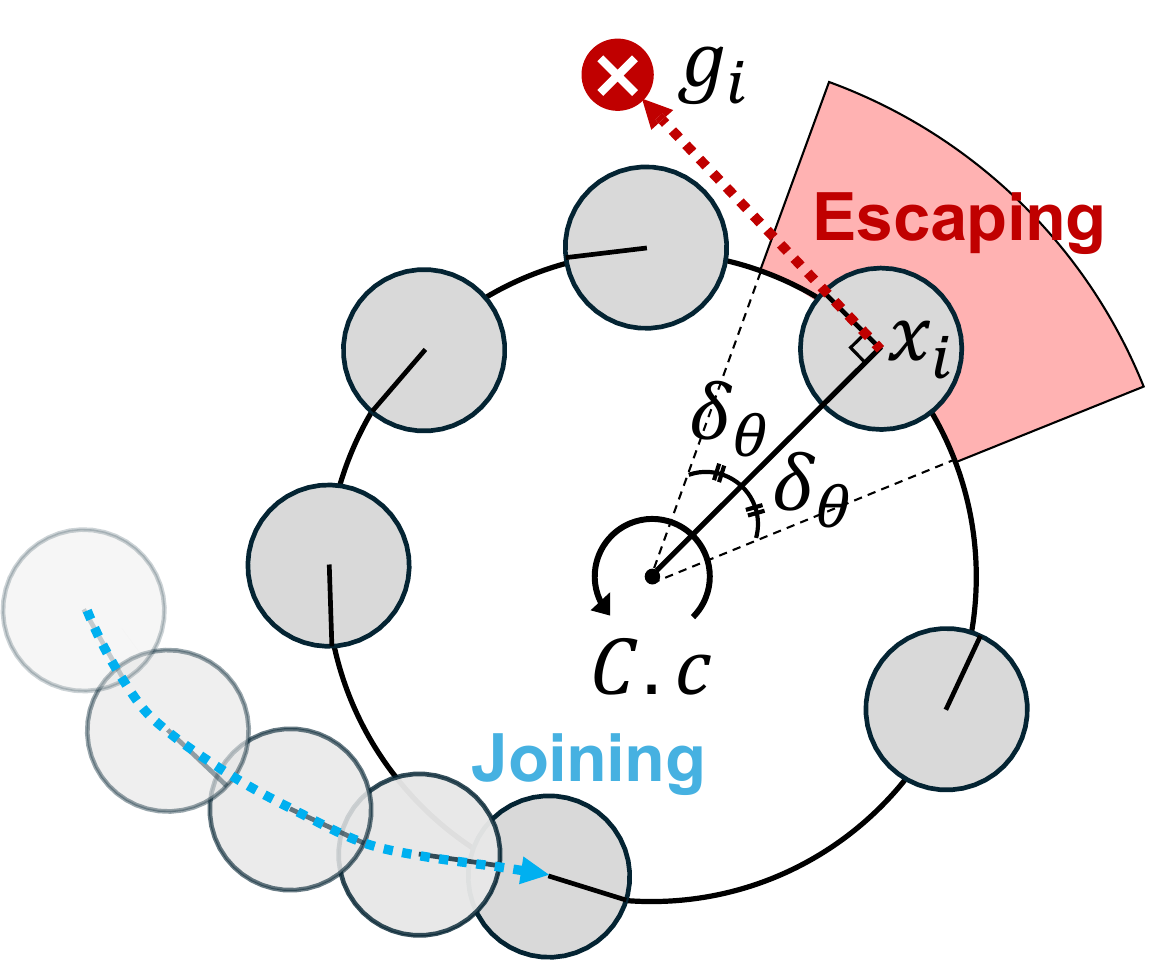}
\vspace{5pt}
\caption{The roundabout joining and escaping mechanism. Some robots joining the circular formation and others escaping when they meet escape conditions. The red shaded sector must be clear of other robots and obstacles for safe escape.}
\label{fig:joining}
\vspace{-10pt}
\end{figure}

Once a valid roundabout $C$ is established, $a_i$ joins $C$ and its mode sets to \textsf{MGR} by \textsc{joinMGR} (line~\ref{alg1:joinMGR}). Since the radial component of (\ref{eq:mgr}) acts proportionally to the error between the current position and the target radius, the robot naturally moves along a spiral approach path when joining the roundabout, as depicted in Fig.~\ref{fig:joining}.
Simultaneously, it broadcasts $C$ to robots within $\delta_\text{comm}$ (i.e., $a_j \in \mathcal{A}$) through \textsc{sendMGR} (line~\ref{alg1:sendMGR}), enabling $a_j$ to join the same avoidance maneuver without requiring centralized coordination. If no nearby roundabout is found, a valid $C$ is created (line~\ref{alg1:createMGR}). The minimum value of the radius $C.r$ is $2r_\text{safe}$ to ensure that the robots taking $C$ do not collide with each other. A large value of $C.r$ is likely to lead to the generation of an invalid roundabout which overlaps with $O$.

In each control period, line~\ref{alg1:isEscapableandaiMGR} checks whether a robot $a_i$ in \textsf{MGR} mode satisfies an escape condition, upon which $a_i$ switches its navigation mode to \textsf{GOAL} (line~\ref{alg1:escape}).
We formalize this condition based on the geometry among the roundabout center $C.c$, the robot position $x_i$, and its goal $g_i$. Let $\mathbf{v}_{ic} = C.c - x_i$ and $\mathbf{v}_{ig} = g_i - x_i$.  If $\mathbf{v}_{ic}$ is orthogonal to $\mathbf{v}_{ig}$, then $a_i$ meets the geometric requirement for escape.  However, $a_i$ must also check the outer region (shown as the red shaded area in Fig.~\ref{fig:joining}) is free from obstacles and other robots. This sector, centered at $C.c$, spans an angle $2\delta_\theta$ and extends to $\|C.c - x_i\| + \delta_\text{sensing}$. If this sector is free of obstacles or robots, $a_i$ can safely escape $C$.

\subsection{Time Complexity Analysis}
We provide a proof for the time complexity of Alg.~\ref{alg:MGR}. 

\thm \textbf{4.1.}
For $N$ robots, the computational complexity of Alg.~\ref{alg:MGR} in each control period is 
\(
O\bigl(N^2 + N\,d_W\,d_H\bigr).
\)\footnote{Here, $d_W$ and $d_H$ are the horizontal and vertical dimensions of the discretized map.}

\begin{proof}
Communication within $\delta_\text{comm}$ is assumed to be $O(1)$. We do not include the time for solving the QP, as our deadlock-prevention logic is layered on top of the controller.

In lines~\ref{alg1:for}--\ref{alg1:endfor}, the for loop iterates up to $N-1$ times, once per neighboring robot. Within each iteration, \textsc{isGoalChecking} checks up to $N$ robots for their goal conditions and modes, contributing $O(N)$. Finding and creating a roundabout center via \textsc{findCenter} and \textsc{createMGR} is constant-time, since these are closed-form operations. Checking validity with \textsc{isMGRValid} also involves a small, fixed set  of geometric comparisons. Updating the navigation mode a robot in \textsc{joinMGR} or \textsc{escapeMGR} is $O(1)$.

In the worst case, \textsc{adjustMGR} searches the entire discretized map of size $d_W\times d_H$, 
yielding $O(d_W\,d_H)$. Finally, \textsc{isEscapable} may need to check up to $N-1$ robots to ensure no one 
blocks the outward path of escaping robot, adding $O(N)$.

Overall, we have $O\bigl(N\,(N + d_W\,d_H) + N\bigr) \;=\; O\bigl(N^2 + N\,d_W\,d_H\bigr).$
\end{proof}

Practically, \textsc{adjustMGR} does not need to scan the entire environment; in our experiments, searching a radius of $10\,C.r$ (i.e., about 3m in a $16\text{m} \times 16\text{m}$ map) around $C.c$ proved sufficient.

\section{Experiments}
We evaluate the proposed algorithm in simulations and also with physical robots. In all experiments, algorithms run on a laptop with an AMD Ryzen 7 5800 8-Core Processor and 32GB RAM, using CVXOPT as the quadratic programming solver. The physical robot experiment is done with DJI RoboMaster S1 robots where their states are measured using a motion capture system (six NOKOV Pluto 1.3C).

\subsection{Setup for Algorithm Tests}

\begin{figure}[t]
\captionsetup{skip=0pt}
\centering
\includegraphics[width=0.99\columnwidth]{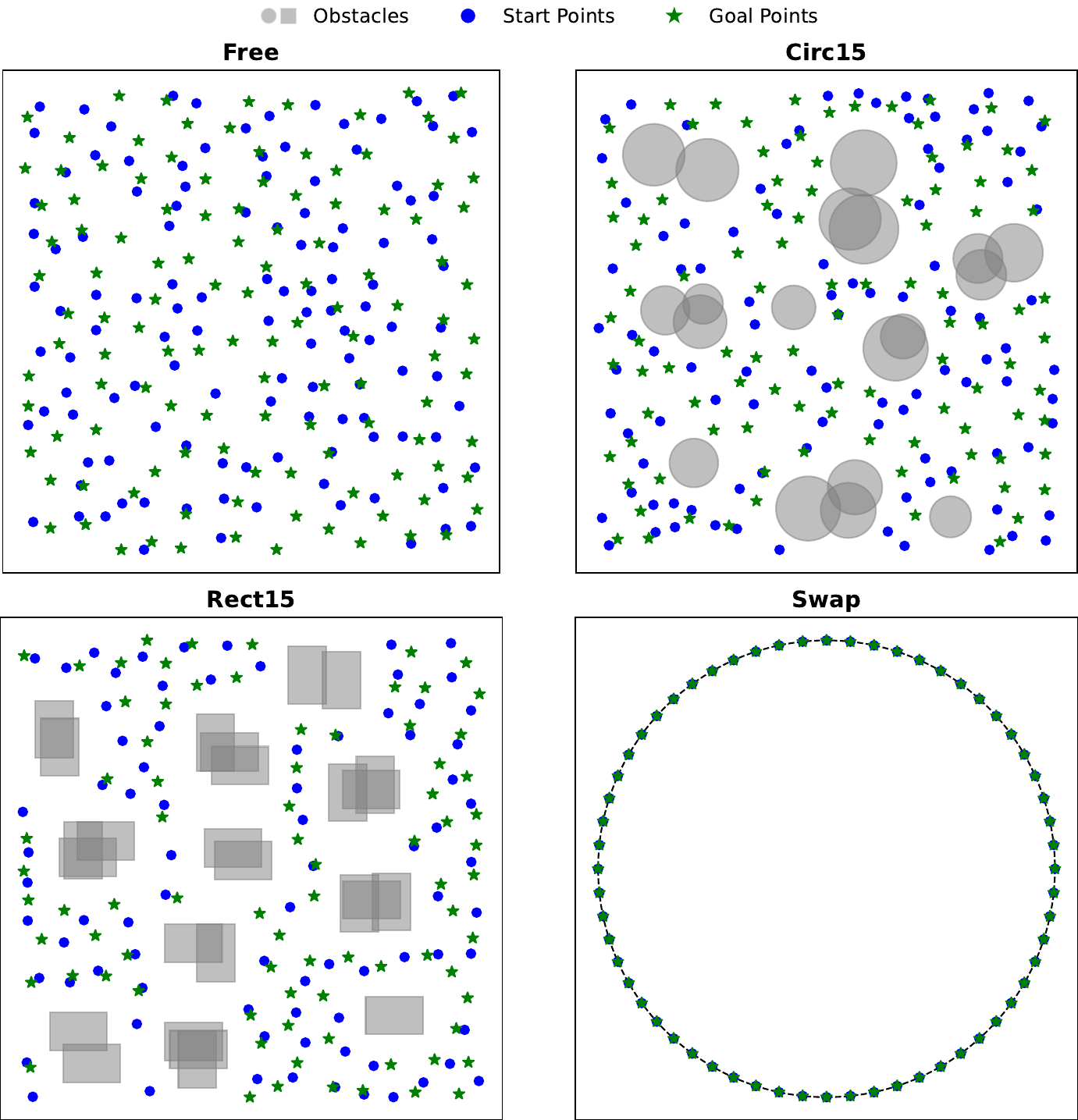}
\caption{The four environments for algorithm tests. (Clockwise from the top-left) They are \textsf{Free} (120 robots and their goals are plotted), \textsf{Circ15} (100 robots),  \textsf{Swap} (60 robots), and \textsf{Rect15} (80 robots).}
\label{fig:env}
\end{figure}

We construct a simulation environment where $W$ is a $16\text{m} \times 16\text{m}$ area. Each robot has a radius of $0.2\text{m}$, and the safe distance $r_{\text{safe}}$ is $0.22\text{m}$ for enforcing barrier constraints. The maximum linear velocity of each robot is 0.8m/s, and its maximum angular velocity is $\pi/2$\,rad/s. We test four different environments, as illustrated in Fig.~\ref{fig:env}. \textsf{Free} is an open space with no obstacles. 
\textsf{Circ15} and \textsf{Rect15} each contain obstacles that occupy 15\% of $W$,  circular in \textsf{Circ15} and rectangular in \textsf{Rect15}.  In these settings, the robot start and goal locations, as well as obstacles, 
are randomly generated for each instance. We generate 20 instances for each value of $N$ (which is up to 120), with all methods evaluated on the same random instances for fair comparison. \textsf{Swap} employs the same obstacle-free environment as \textsf{Free}, but places the robots 15m from the center on opposite sides, 
requiring them to swap positions.

We compare our method with three widely used decentralized control techniques: GCBF+~\cite{zhang2025gcbf+}, ORCA~\cite{van2011reciprocal}, and CLF-CBF~\cite{ames2017control}. We use four metrics: (i)~\emph{success rate}, counting an instance as successful only if \emph{all} robots reach their goals within a 2-minute limit; instances exceeding 2 minutes typically stall due to deadlock; (ii)~\emph{arrival rate}, the proportion of robots that reach their goals; (iii)~\emph{makespan}, the time until all robots have arrived in a successful instance (makespan is undefined in failed instances); and (iv)~\emph{mean time}, the average time among successfully arriving robots. Since non-arrivals are excluded, this metric can be skewed if many robots fail. If a method is with low arrival rates, the mean time only includes the result for relatively easy goals and simply discards challenging cases, thus artificially reducing their reported means.

For all experiments, we implement our QP formulation~(\ref{eq:qp_obj}) with a positive definite Hessian matrix $H = \text{diag}(2, 2, 1)$. The linear term is chosen as $F = [-2(u_i^{\text{des}})^T, 0]^T$ for $a_i$.
The CLF is implemented as a quadratic function $V(x_i)=(x_i-g_i)^TP(x_i-g_i)$ with $P=diag(1,1)$, and $\gamma(V)=\lambda V$ where $\lambda =1$. The CBF constraints are constructed using functions with $\alpha(h)=\beta h$ where $\beta=5$. In addition, we use the following parameter values: deadlock prediction threshold $k_D = 1$, roundabout proximity $\delta_c = 2$m, roundabout radius $C.r = 0.3$m, communication range $\delta_{\text{comm}} = 1$m, radius increment constant $k = 0.1$m, and proportional gain $k_p = 0.05$. For the escape angle threshold $\delta_{\theta}$, we used different values depending on the environment: $\delta_{\theta} = \pi/6$ for environments with obstacles and $\delta_\theta = \pi/12$ for obstacle-free environments.

\subsection{Results}
\begin{figure}[t]
\captionsetup{skip=0pt}
\centering
\includegraphics[width=0.98\columnwidth]{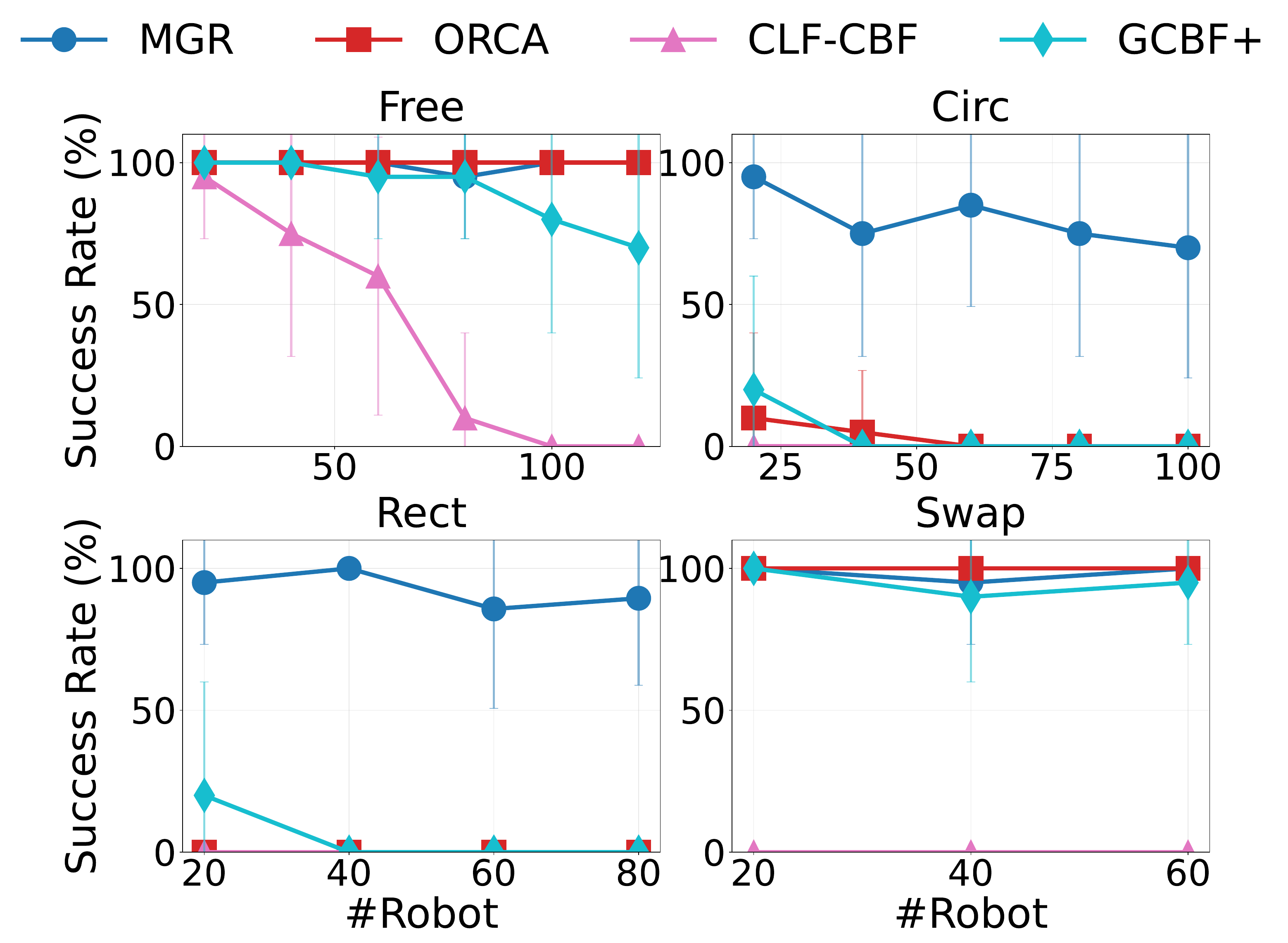}
\caption{The success rates of the compared methods}
\label{fig:success}\label{fig:rates}
\vspace{-10pt}
\end{figure}

\begin{figure}[t]
\captionsetup{skip=0pt}
\centering
\includegraphics[width=0.98\columnwidth]{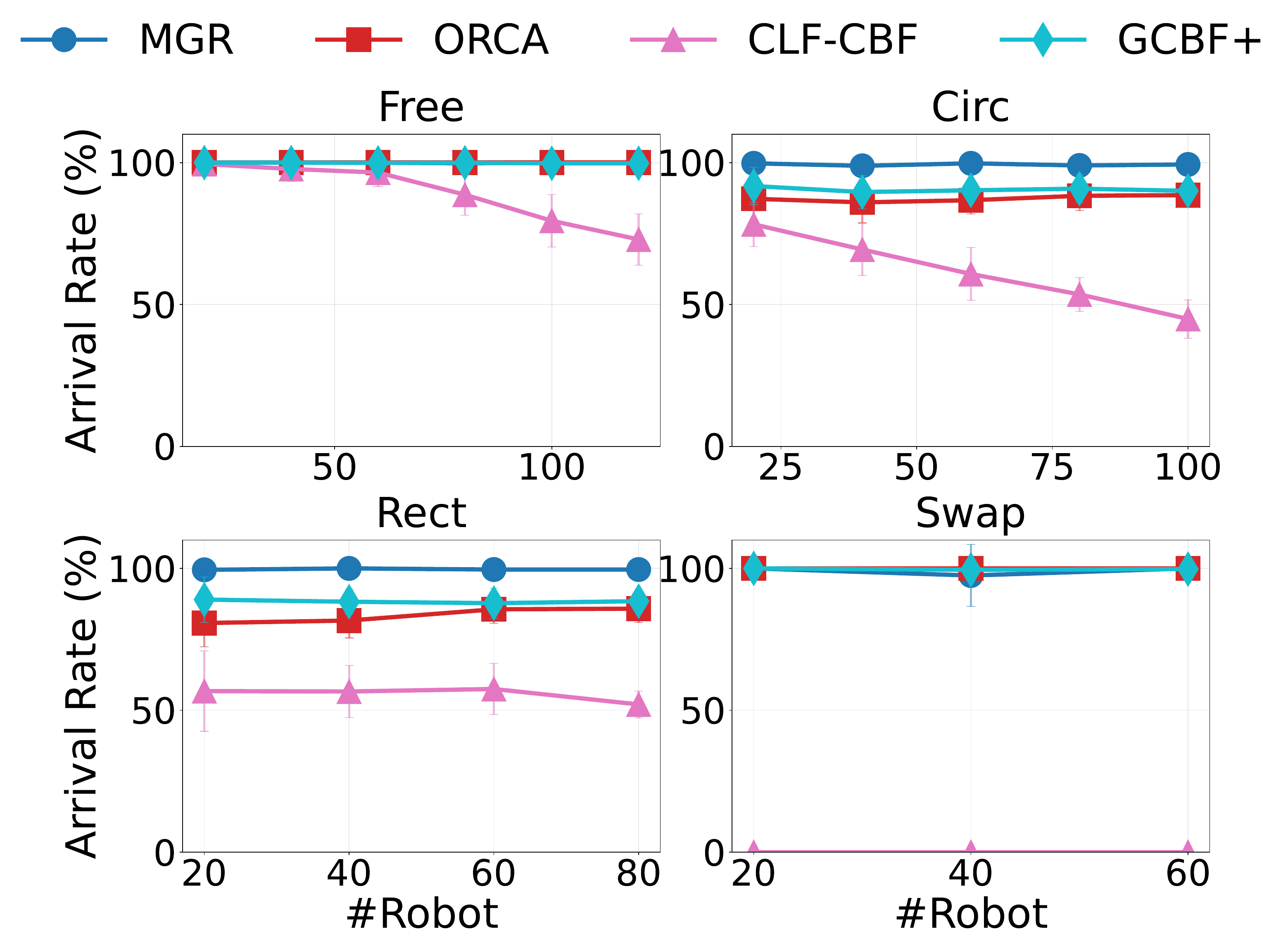}
\caption{The arrival rates of the compared methods}
\label{fig:arrival}
\vspace{-5pt}
\end{figure}

Success and arrival results are summarized in Figs.~\ref{fig:success} and \ref{fig:arrival} and Table~\ref{tab:rates}. Overall, our proposed method (MGR) consistently achieves superior or comparable performance across all environments and robot densities. In the obstacle-free setting (\textsf{Free}), MGR maintains near-perfect arrival and success rates even as the number of robots increases, outperforming CLF-CBF and  GCBF+ and keeping pace with ORCA. In more difficult environments (\textsf{Circ15} and \textsf{Rect15}), MGR clearly outperforms the other methods: its arrival rates remain above 98\%, and, crucially, it achieves a nonzero success rate for high robot counts—whereas most other methods drop to 0\% success. Finally, in the \textsf{Swap} scenario, the performance of MGR remains robust, with success rates ranging from 95\% to 100\%, closely matching ORCA and surpassing CLF-CBF outright. This trend highlights the effectiveness MGR in both open and obstacle-dense environments, especially under high robot densities, where it maintains stable success and arrival rates that other methods struggle to achieve.

\begin{table*}[]
\centering
\captionsetup{skip=0pt}
\caption{The success and arrival rates calculated from 20 instances for each team size and environment}\label{tab:rates}
\resizebox{0.67\textwidth}{!}{
\begin{tabular}{|c|c|cccc|cccc|}
\hline
\multirow{2}{*}{Env.}                       & \multirow{2}{*}{\#robot} & \multicolumn{4}{c|}{Success Rate (\%)}                                                            & \multicolumn{4}{c|}{Arrival Rate (\%)}                                                            \\ \cline{3-10} 
                                            &                          & \multicolumn{1}{c|}{MGR}    & \multicolumn{1}{c|}{ORCA}   & \multicolumn{1}{c|}{CLF-CBF} & GCBF+  & \multicolumn{1}{c|}{MGR}    & \multicolumn{1}{c|}{ORCA}   & \multicolumn{1}{c|}{CLF-CBF} & GCBF+  \\ \hline
\multirow{6}{*}{\textsf{Free}} & 20                       & \multicolumn{1}{c|}{100.00} & \multicolumn{1}{c|}{100.00} & \multicolumn{1}{c|}{95.00}   & 100.00 & \multicolumn{1}{c|}{100.00} & \multicolumn{1}{c|}{100.00} & \multicolumn{1}{c|}{99.49}   & 100.00 \\ \cline{2-10} 
                                            & 40                       & \multicolumn{1}{c|}{100.00} & \multicolumn{1}{c|}{100.00} & \multicolumn{1}{c|}{75.00}   & 100.00 & \multicolumn{1}{c|}{100.00} & \multicolumn{1}{c|}{100.00} & \multicolumn{1}{c|}{97.75}   & 100.00 \\ \cline{2-10} 
                                            & 60                       & \multicolumn{1}{c|}{100.00} & \multicolumn{1}{c|}{100.00} & \multicolumn{1}{c|}{60.00}   & 95.00  & \multicolumn{1}{c|}{100.00} & \multicolumn{1}{c|}{100.00} & \multicolumn{1}{c|}{96.50}   & 99.91  \\ \cline{2-10} 
                                            & 80                       & \multicolumn{1}{c|}{95.00}  & \multicolumn{1}{c|}{100.00} & \multicolumn{1}{c|}{10.00}   & 95.00  & \multicolumn{1}{c|}{99.75}  & \multicolumn{1}{c|}{100.00} & \multicolumn{1}{c|}{88.68}   & 99.93  \\ \cline{2-10} 
                                            & 100                      & \multicolumn{1}{c|}{100.00} & \multicolumn{1}{c|}{100.00} & \multicolumn{1}{c|}{0.00}    & 80.00  & \multicolumn{1}{c|}{100.00} & \multicolumn{1}{c|}{100.00} & \multicolumn{1}{c|}{79.50}   & 99.75  \\ \cline{2-10} 
                                            & 120                      & \multicolumn{1}{c|}{100.00} & \multicolumn{1}{c|}{100.00} & \multicolumn{1}{c|}{0.00}    & 70.00  & \multicolumn{1}{c|}{100.00} & \multicolumn{1}{c|}{100.00} & \multicolumn{1}{c|}{72.91}   & 99.75  \\ \hline
\multirow{5}{*}{\textsf{Circ15}} & 20                       & \multicolumn{1}{c|}{95.00}  & \multicolumn{1}{c|}{10.00}  & \multicolumn{1}{c|}{0.00}    & 20.00  & \multicolumn{1}{c|}{99.75}  & \multicolumn{1}{c|}{87.25}  & \multicolumn{1}{c|}{78.25}   & 91.74  \\ \cline{2-10} 
                                            & 40                       & \multicolumn{1}{c|}{75.00}  & \multicolumn{1}{c|}{5.00}   & \multicolumn{1}{c|}{0.00}    & 0.00   & \multicolumn{1}{c|}{98.87}  & \multicolumn{1}{c|}{86.00}  & \multicolumn{1}{c|}{69.37}   & 89.62  \\ \cline{2-10} 
                                            & 60                       & \multicolumn{1}{c|}{85.00}  & \multicolumn{1}{c|}{0.00}   & \multicolumn{1}{c|}{0.00}    & 0.00   & \multicolumn{1}{c|}{99.75}  & \multicolumn{1}{c|}{86.75}  & \multicolumn{1}{c|}{60.75}   & 90.25  \\ \cline{2-10} 
                                            & 80                       & \multicolumn{1}{c|}{75.00}  & \multicolumn{1}{c|}{0.00}   & \multicolumn{1}{c|}{0.00}    & 0.00   & \multicolumn{1}{c|}{99.00}  & \multicolumn{1}{c|}{88.31}  & \multicolumn{1}{c|}{53.56}   & 90.81  \\ \cline{2-10} 
                                            & 100                      & \multicolumn{1}{c|}{70.00}  & \multicolumn{1}{c|}{0.00}   & \multicolumn{1}{c|}{0.00}    & 0.00   & \multicolumn{1}{c|}{99.35}  & \multicolumn{1}{c|}{88.55}  & \multicolumn{1}{c|}{44.90}   & 90.05  \\ \hline
\multirow{4}{*}{\textsf{Rect15}} & 20                       & \multicolumn{1}{c|}{95.00}  & \multicolumn{1}{c|}{0.00}   & \multicolumn{1}{c|}{0.00}    & 20.00  & \multicolumn{1}{c|}{99.49}  & \multicolumn{1}{c|}{80.75}  & \multicolumn{1}{c|}{56.75}   & 89.00  \\ \cline{2-10} 
                                            & 40                       & \multicolumn{1}{c|}{100.00} & \multicolumn{1}{c|}{0.00}   & \multicolumn{1}{c|}{0.00}    & 0.00   & \multicolumn{1}{c|}{100.00} & \multicolumn{1}{c|}{81.62}  & \multicolumn{1}{c|}{56.62}   & 88.25  \\ \cline{2-10} 
                                            & 60                       & \multicolumn{1}{c|}{85.00}  & \multicolumn{1}{c|}{0.00}   & \multicolumn{1}{c|}{0.00}    & 0.00   & \multicolumn{1}{c|}{99.60}  & \multicolumn{1}{c|}{85.58}  & \multicolumn{1}{c|}{57.50}   & 87.75  \\ \cline{2-10} 
                                            & 80                       & \multicolumn{1}{c|}{90.00}  & \multicolumn{1}{c|}{0.00}   & \multicolumn{1}{c|}{0.00}    & 0.00   & \multicolumn{1}{c|}{99.60}  & \multicolumn{1}{c|}{85.81}  & \multicolumn{1}{c|}{52.06}   & 88.43  \\ \hline
\multirow{3}{*}{\textsf{Swap}} & 20                       & \multicolumn{1}{c|}{100.00} & \multicolumn{1}{c|}{100.00} & \multicolumn{1}{c|}{0.00}    & 100.00 & \multicolumn{1}{c|}{100.00} & \multicolumn{1}{c|}{100.00} & \multicolumn{1}{c|}{0.00}    & 100.00 \\ \cline{2-10} 
                                            & 40                       & \multicolumn{1}{c|}{95.00}  & \multicolumn{1}{c|}{100.00} & \multicolumn{1}{c|}{0.00}    & 90.00  & \multicolumn{1}{c|}{97.50}  & \multicolumn{1}{c|}{100.00} & \multicolumn{1}{c|}{0.00}    & 99.49  \\ \cline{2-10} 
                                            & 60                       & \multicolumn{1}{c|}{100.00} & \multicolumn{1}{c|}{100.00} & \multicolumn{1}{c|}{0.00}    & 95.00  & \multicolumn{1}{c|}{100.00} & \multicolumn{1}{c|}{100.00} & \multicolumn{1}{c|}{0.00}    & 99.75  \\ \hline
\end{tabular}
}
\end{table*}

The makespan and mean navigation time are shown in Table~\ref{tab:times}. MGR generally reports longer makespans and mean times --- reflecting its more conservative roundabout-based approach --- yet it consistently completes all scenarios, even with large teams or dense obstacles. In contrast, ORCA, GCBF+, and especially CLF-CBF often fail or yield infeasible times (shown as ``–'') under higher robot counts or  cluttered layouts. This trade-off demonstrates that the extra caution of MGR pays off in higher success rates and reliable arrivals, making it more robust in challenging conditions where other methods struggle or stall.

We also attempted to compare with NMPC~\cite{lafmejani2021nonlinear}, which explicitly addresses deadlock prevention. However, NMPC becomes computationally prohibitive even at moderate team sizes: 
with just 20 robots, it requires over 10 seconds per robot for each control step,  ruling out real-time implementation. In contrast, our approach remains efficient across all tests, never exceeding 2ms per robot per control period,  even with 120 robots. This result indicates that our deadlock prevention strategy 
adds negligible overhead while still reliably avoiding deadlocks.

\begin{table*}[]
\centering
\captionsetup{skip=0pt}
\caption{The makespan and mean time where the numbers in parenthesis are standard deviations}\label{tab:times}
\resizebox{0.88\textwidth}{!}{
\begin{tabular}{|c|c|cccc|cccc|}
\hline
\multirow{2}{*}{Env.}                         & \multirow{2}{*}{\#robot} & \multicolumn{4}{c|}{Makespan (sec)}                                                                                         & \multicolumn{4}{c|}{Mean time (sec)}                                                                                                          \\ \cline{3-10} 
&                          & \multicolumn{1}{c|}{MGR}           & \multicolumn{1}{c|}{ORCA}          & \multicolumn{1}{c|}{CLF-CBF}       & GCBF+        & \multicolumn{1}{c|}{MGR}          & \multicolumn{1}{c|}{ORCA}         & \multicolumn{1}{c|}{CLF-CBF}      & GCBF+                             \\ \hline
\multirow{6}{*}{\textsf{Free}}   & 20                       & \multicolumn{1}{c|}{21.19 (2.93)}  & \multicolumn{1}{c|}{18.38 (2.45)}  & \multicolumn{1}{c|}{20.32 (2.49)}  & 31.37 (4.36) & \multicolumn{1}{c|}{12.79 (1.62)} & \multicolumn{1}{c|}{9.97 (1.11)}  & \multicolumn{1}{c|}{11.59 (1.21)} & \multicolumn{1}{l|}{16.53 (1.91)} \\ \cline{2-10} 
& 40                       & \multicolumn{1}{c|}{26.10 (3.01)}  & \multicolumn{1}{c|}{19.74 (1.88)}  & \multicolumn{1}{c|}{23.00 (2.36)}  & 34.22 (3.40) & \multicolumn{1}{c|}{14.89 (1.30)} & \multicolumn{1}{c|}{9.86 (0.73)}  & \multicolumn{1}{c|}{11.86 (0.78)} & \multicolumn{1}{l|}{16.05 (1.28)} \\ \cline{2-10} 
& 60                       & \multicolumn{1}{c|}{32.15 (5.57)}  & \multicolumn{1}{c|}{22.31 (2.67)}  & \multicolumn{1}{c|}{33.69 (26.33)} & 36.48 (4.24) & \multicolumn{1}{c|}{18.04 (1.99)} & \multicolumn{1}{c|}{10.74 (0.80)} & \multicolumn{1}{c|}{13.10 (1.18)} & \multicolumn{1}{l|}{16.86 (0.93)} \\ \cline{2-10} 
& 80                       & \multicolumn{1}{c|}{44.17 (11.03)} & \multicolumn{1}{c|}{22.93 (2.07)}  & \multicolumn{1}{c|}{28.05 (0.30)}  & 36.59 (2.79) & \multicolumn{1}{c|}{22.07 (1.57)} & \multicolumn{1}{c|}{11.01 (0.55)} & \multicolumn{1}{c|}{13.38 (1.23)} & \multicolumn{1}{l|}{17.13 (1.08)} \\ \cline{2-10} 
& 100                      & \multicolumn{1}{c|}{53.35 (12.15)} & \multicolumn{1}{c|}{25.73 (4.19)}  & \multicolumn{1}{c|}{-}             & 39.77 (3.45) & \multicolumn{1}{c|}{27.70 (2.80)} & \multicolumn{1}{c|}{11.80 (0.79)} & \multicolumn{1}{c|}{14.05 (1.37)} & \multicolumn{1}{l|}{17.51 (0.87)} \\ \cline{2-10} 
& 120                      & \multicolumn{1}{c|}{60.44 (10.85)} & \multicolumn{1}{c|}{28.98 (3.09)}  & \multicolumn{1}{c|}{-}             & 40.81 (3.67) & \multicolumn{1}{c|}{33.29 (3.21)} & \multicolumn{1}{c|}{12.65 (0.85)} & \multicolumn{1}{c|}{14.54 (1.19)} & \multicolumn{1}{l|}{17.76 (0.79)} \\ \hline
\multirow{5}{*}{\textsf{Circ15}} & 20                       & \multicolumn{1}{c|}{31.68 (7.06)}  & \multicolumn{1}{c|}{49.90 (0.85)}  & \multicolumn{1}{c|}{-}             & 33.93 (3.81) & \multicolumn{1}{c|}{17.35 (3.10)} & \multicolumn{1}{c|}{13.14 (2.64)} & \multicolumn{1}{c|}{12.38 (2.40)} & 18.47 (3.75)                      \\ \cline{2-10} 
& 40                       & \multicolumn{1}{c|}{39.93 (8.05)}  & \multicolumn{1}{c|}{100.60 (0.00)} & \multicolumn{1}{c|}{-}             & -            & \multicolumn{1}{c|}{21.02 (2.65)} & \multicolumn{1}{c|}{15.14 (2.63)} & \multicolumn{1}{c|}{12.19 (1.26)} & 18.38 (1.56)                      \\ \cline{2-10} 
& 60                       & \multicolumn{1}{c|}{53.22 (7.88)}  & \multicolumn{1}{c|}{-}             & \multicolumn{1}{c|}{-}             & -            & \multicolumn{1}{c|}{27.54 (2.50)} & \multicolumn{1}{c|}{15.09 (1.71)} & \multicolumn{1}{c|}{12.73 (1.71)} & 19.52 (1.37)                      \\ \cline{2-10} 
& 80                       & \multicolumn{1}{c|}{67.76 (12.69)} & \multicolumn{1}{c|}{-}             & \multicolumn{1}{c|}{-}             & -            & \multicolumn{1}{c|}{33.14 (3.39)} & \multicolumn{1}{c|}{17.49 (2.38)} & \multicolumn{1}{c|}{13.06 (1.90)} & 18.80 (1.22)                      \\ \cline{2-10} 
& 100                      & \multicolumn{1}{c|}{81.11 (12.59)} & \multicolumn{1}{c|}{-}             & \multicolumn{1}{c|}{-}             & -            & \multicolumn{1}{c|}{42.57 (5.31)} & \multicolumn{1}{c|}{17.48 (2.37)} & \multicolumn{1}{c|}{12.95 (1.56)} & 19.62 (1.22)                      \\ \hline
\multirow{4}{*}{\textsf{Rect15}} & 20                       & \multicolumn{1}{c|}{37.63 (11.00)} & \multicolumn{1}{c|}{-}             & \multicolumn{1}{c|}{-}             & 41.87 (8.25) & \multicolumn{1}{c|}{19.15 (3.44)} & \multicolumn{1}{c|}{13.23 (2.69)} & \multicolumn{1}{c|}{11.59 (1.12)} & 19.09 (2.60)                      \\ \cline{2-10} 
& 40                       & \multicolumn{1}{c|}{46.30 (8.81)}  & \multicolumn{1}{c|}{-}             & \multicolumn{1}{c|}{-}             & -            & \multicolumn{1}{c|}{23.99 (2.86)} & \multicolumn{1}{c|}{13.65 (1.58)} & \multicolumn{1}{c|}{12.30 (2.06)} & 19.44 (2.14)                      \\ \cline{2-10} 
& 60                       & \multicolumn{1}{c|}{57.17 (10.44)} & \multicolumn{1}{c|}{-}             & \multicolumn{1}{c|}{-}             & -            & \multicolumn{1}{c|}{31.00 (4.46)} & \multicolumn{1}{c|}{15.47 (2.18)} & \multicolumn{1}{c|}{13.02 (1.67)} & 19.21 (1.50)                      \\ \cline{2-10} 
& 80                       & \multicolumn{1}{c|}{69.90 (13.48)} & \multicolumn{1}{c|}{-}             & \multicolumn{1}{c|}{-}             & -            & \multicolumn{1}{c|}{39.11 (4.19)} & \multicolumn{1}{c|}{15.50 (1.68)} & \multicolumn{1}{c|}{12.86 (1.74)} & 19.15 (1.17)                      \\ \hline
\multirow{3}{*}{\textsf{Swap}}   & 20                       & \multicolumn{1}{c|}{37.67 (4.45)}  & \multicolumn{1}{c|}{32.15 (0.00)}  & \multicolumn{1}{c|}{-}             & 36.84 (0.00) & \multicolumn{1}{c|}{30.38 (2.10)} & \multicolumn{1}{c|}{26.93 (0.00)} & \multicolumn{1}{c|}{-}            & 33.72 (0.03)                      \\ \cline{2-10} 
& 40                       & \multicolumn{1}{c|}{44.77 (10.12)} & \multicolumn{1}{c|}{42.35 (0.00)}  & \multicolumn{1}{c|}{-}             & 41.8 (5.59)  & \multicolumn{1}{c|}{33.96 (6.36)} & \multicolumn{1}{c|}{34.82 (0.00)} & \multicolumn{1}{c|}{-}            & 35.63 (0.95)                      \\ \cline{2-10} 
& 60                       & \multicolumn{1}{c|}{52.61 (13.31)} & \multicolumn{1}{c|}{45.10 (0.00)}  & \multicolumn{1}{c|}{-}             & 44.39 (3.32) & \multicolumn{1}{c|}{36.22 (3.39)} & \multicolumn{1}{c|}{35.63 (0.00)} & \multicolumn{1}{c|}{-}            & 36.52 (0.36)                      \\ \hline
\end{tabular}
}
\end{table*}

\subsection{Physical Robot Experiment}
For our physical robot experiments, we use a 3m $\times$ 3.5m bounded workspace with robots (dimension is 320mm $\times$ 240mm). We run our method in three scenarios: \textsf{Free}, \textsf{Swap}, and \textsf{Rect}. The maximum linear velocity of each robot is 0.5m/s, and its maximum angular velocity is $\pi/2$\,rad/s. Although the physical robots support omnidirectional motion (including lateral movement), we restrict them to differential drive controls for consistency with our setup. A summary of the results can be found in the supplementary material.

\section{Conclusion}
\vspace{-2pt}

We proposed Merry-Go-Round (MGR), a decentralized multi-robot navigation framework that augments a standard controller with explicit deadlock prevention. By detecting deadlocks, robots form local roundabouts and move in a controlled orbit until safe to proceed. This lightweight, peer-to-peer approach is highly scalable. Extensive simulations and physical robot tests show that MGR consistently achieves high success and arrival rates, outperforming or equaling existing decentralized controllers, especially in dense or cluttered environments where deadlocks commonly occur.

\bibliographystyle{IEEEtran}
\bibliography{references}

\end{document}